\documentclass{article}

\usepackage{PRIMEarxiv}

\makeatletter
\renewcommand{\normalsize}{%
  \@setfontsize\normalsize{12pt}{14pt}% <- change these two numbers
  \abovedisplayskip      7\p@ \@plus 2\p@ \@minus 5\p@
  \abovedisplayshortskip \z@ \@plus 3\p@
  \belowdisplayskip      \abovedisplayskip
  \belowdisplayshortskip 4\p@ \@plus 3\p@ \@minus 3\p@
}
\makeatother
\AtBeginDocument{\normalsize}

\usepackage[utf8]{inputenc}
\usepackage[T1]{fontenc}
\usepackage[hidelinks]{hyperref}
\usepackage{url}
\usepackage{microtype}
\usepackage{graphicx}
\usepackage{subcaption}
\usepackage{booktabs}
\usepackage[table]{xcolor}
\usepackage{xspace}
\usepackage{enumitem}
\usepackage{amsmath,amssymb,mathtools,amsthm}
\usepackage{amsfonts}
\usepackage{nicefrac}
\usepackage{algorithm}
\usepackage{algorithmic}
\usepackage{multirow}
\usepackage{multicol}
\usepackage{array}
\usepackage{float}
\usepackage[most]{tcolorbox}
\usepackage{tikz}
\usepackage{listings}
\usepackage{wrapfig}

\graphicspath{{images/}{images/experiments/}}

\usepackage{fancyhdr}
\usepackage{fontawesome5}
\hypersetup{
  colorlinks=true,
  urlcolor=linkblue,
  linkcolor=linkblue,
  citecolor=linkblue
}

\makeatletter
\renewcommand{\@toptitlebar}{\vskip 0.25in\vskip -\parskip}
\renewcommand{\@bottomtitlebar}{\vskip 0.29in\vskip -\parskip\vskip 0.09in}
\makeatother

\definecolor{promptgold}{HTML}{FFF59D}
\definecolor{revmagenta}{HTML}{D62AA0}
\definecolor{framegray}{HTML}{C9CEDA}
\definecolor{promptazure}{HTML}{E6F6FF}

\definecolor{linkblue}{RGB}{0,102,204}

\newcommand{\mask}{\texttt{[MASK]}}
\newcommand{\method}{\textsc{CoRe}\xspace}
\DeclareMathOperator*{\argmin}{arg\,min}

\newcommand{\midscriptsize}{\fontsize{8}{9}\selectfont}
\newcommand{\revise}[1]{\colorbox{magenta!25}{\textcolor{magenta}{\ttfamily #1}}}
\newcommand{\masking}[1]{\colorbox{yellow!25}{\textcolor{black}{\ttfamily #1}}}
\newcommand{\correct}[1]{\colorbox{green!25}{\textcolor{black}{\ttfamily #1}}}

\newtcolorbox{promptline}{
  colback=promptazure,
  colframe=promptazure,
  boxrule=0pt,
  arc=0mm,
  left=3pt,right=3pt,top=2pt,bottom=2pt,
  boxsep=0pt,
  enhanced,
  breakable
}

\theoremstyle{plain}

\theoremstyle{definition}

\theoremstyle{remark}

\usepackage[numbers]{natbib}

\title{\method: Context-Robust Remasking for Diffusion Language Models}

\author{Kevin Zhai \And Sabbir Mollah \And Zhenyi Wang \And Mubarak Shah}

\date{}

\begin{document}
\maketitle

\vspace{-0.45in}
\begin{center}
  \normalsize University of Central Florida \\ [12pt]
  \faGlobe\ \href{https://ucf-crcv.github.io/core/}{Project} \quad
  \faGithub\ \href{https://github.com/UCF-CRCV/core}{Code}
\end{center}
\vspace{0.35in}

\begin{abstract}

Standard decoding in Masked Diffusion Models (MDMs) is hindered by context rigidity: tokens are retained based on transient high confidence, often ignoring that early predictions lack full context. This creates cascade effects where initial inconsistencies misguide the remaining generation. Existing revision strategies attempt to mitigate this by relying on static confidence scores, but these signals are inherently short-sighted; inconsistent tokens frequently appear confident to the model itself. To address this, we propose \textbf{Co}ntext-Robust \textbf{Re}masking (\method), a training-free framework for inference-time revision. Rather than trusting static token probabilities, we identify \textit{context-brittle} tokens by probing their sensitivity to targeted perturbations. We formalize revision as a robust optimization problem targeting worst-case context shifts. \method efficiently approximates this objective to expose unstable tokens, prioritizing them for revision. On LLaDA-8B-Base, \method delivers consistent improvements across reasoning and code benchmarks, outperforming compute-matched baselines and boosting performance on code generation (MBPP) by up to $+9.2\%$.

\end{abstract}

\section{Introduction}

Masked Diffusion Models (MDMs) have emerged as a promising alternative to autoregressive decoding for discrete sequence modeling~\cite{sahoo2024mdlm}, allowing parallel token updates through iterative unmasking~\cite{nie2025large}. At inference time, an MDM starts with a fully masked sequence and progressively constructs a response. At each diffusion step, the sampler predicts token distributions for masked positions conditioned on the current partially unmasked sequence, selects a subset of positions to unmask, and unmasks them. This iterative process means that unmasked tokens in the early steps are selected under a partial context that evolves as decoding proceeds. To improve quality, some variants employ revision strategies, remasking a small subset of previously unmasked tokens to allow resampling~\cite{wang2025remasking}.

Many existing training-free remasking strategies rely on heuristic selection rules, targeting tokens with low confidence (low probability) or small top-$k$ margins (the gap between the top two candidate tokens). These scores are typically computed under a \textit{single} partially unmasked state—or recorded at the step when a token is first selected. However, this approach does not account for an important property of diffusion decoding: as the model generates tokens at new positions, the surrounding context is continuously updated. Consequently, a token may receive high probability under an ambiguous early context, yet later become incompatible once additional tokens stabilize the surrounding structure. Hence, confidence-based heuristics often miss context-brittle tokens that appear reliable but are sensitive to context shifts. Reliability, therefore, depends on robustness to context perturbations, not static uncertainty. Overlooking this distinction leads to suboptimal revision decisions and degraded performance, especially on structure-sensitive generation (e.g., code), where early structural commitments can propagate and amplify downstream errors.

To address this limitation, we propose {\bf Co}ntext-Robust {\bf Re}masking (\method). Instead of ranking tokens by static confidence scores, \method measures whether each token is still strongly predicted when parts of its surrounding context are masked. A reliable token should remain strongly predicted under these masked-context perturbations. Concretely, \method performs a lightweight stress test by evaluating tokens under a restricted family of masked-context perturbations and prioritizing those with the largest drop in support (i.e., highest instability) for revision. As a result, the revision in \method is adaptive: instead of relying on stale uncertainty estimates, it targets tokens that are most vulnerable to dynamic context changes as decoding progresses. This procedure enables \method to remask the most context-sensitive tokens.

Our method improves masked diffusion decoding within a fixed inference budget. We observe the greatest gains on structure-sensitive tasks: \method increases MBPP accuracy by up to \textbf{+9.2\%}, reducing syntax and logic inconsistencies where baselines fail. In contrast, we find that standard confidence-based remasking strategies (e.g., ReMDM~\citep{wang2025remasking}) can degrade code performance in our experiments. These results support our central claim: intelligently allocating a small fraction of compute to stress-test context-brittle tokens is more effective than standard confidence-based revision. This finding is further validated by compute-matched controls, where random or margin-based revision yields negligible gains.

Our contributions are as follows.
\begin{itemize}[leftmargin=0.5cm,itemsep=6pt]
    \item We introduce \method, a training-free framework for remasking in diffusion language models. \method selects revision targets based on robustness to perturbations of the conditioning context, rather than token-specific uncertainty heuristics (e.g., confidence) computed under a single decoding state.
    \item We develop an efficient decoding algorithm that implements this framework via a lightweight stress test. This provides a tractable proxy for our worst-case instability objective by simultaneously masking a candidate subset of positions in a single forward pass, enabling context-aware remasking with minimal overhead.
    \item We demonstrate that \method achieves superior generation quality compared to baselines at equivalent inference cost, with the largest gains on structure-sensitive code generation (up to \textbf{+9.2\%} on MBPP) and consistent performance on reasoning benchmarks.
\end{itemize}

\section{Related Work}

Non-autoregressive generation has evolved from iterative editing objectives~\citep{gu2018nat,stern2019insertion,ghazvininejad2019mask,levenshtein,stern2019insertion,chang2022maskgit} to discrete diffusion models. Early works, such as D3PM~\cite{austin2023d3pm,sohl2015deep,ho2020denoising} and SEDD~\cite{lou2024sedd}, formulate text generation as a discrete state-space diffusion process. MDLM~\cite{sahoo2024mdlm} simplifies this by treating generation as a masked language modeling objective, while LLaDA~\cite{nie2025large} and successors~\citep{ye2025dream,arriola2025block,li2025refusion,shi2024simplified,ye2024diffusion} scale this approach to billions of parameters. Recent works have further extended this paradigm to support variable-length generation, such as FlexMDM~\citep{kim2025anyorder}, or captured latent dependencies via variational lower bounds~\cite{zhang2025variational}. Despite these modeling advances, standard unmasking strategies remain inflexible: they lack a mechanism to revisit decisions once a token is unmasked. To enable revision, ReMDM~\cite{wang2025remasking} introduces step-dependent remasking, while P2-Self~\cite{peng2025p2self} separates decoding into planning and generation.

Crucially, these revision methods either rely on unreliable proxy quality signals or require additional training. ReMDM's standard strategy uses stale confidence scores from past decoding steps, ignoring updated context. P2-Self uses current token probabilities, which can be misleadingly high even for coherent but incorrect generations. Similarly, recent concurrent works like {Deferred Commitment Decoding (DCD)}~\citep{shu2026deferred}, {Coherent Contextual Decoding (CCD)}~\citep{chen2025beyond}, or {Information-Gain (IG) Sampling}~\citep{yang2026improvingsamplingmaskeddiffusion} attempt to mitigate over-confidence by optimizing unmasking order or dynamically delaying commitment. While IG Sampling attempts to minimize future uncertainty by maximizing information gain during selection, it remains a forward-only heuristic. These strategies remain fundamentally passive: they rely on unperturbed internal states and lack a mechanism to revisit previously unmasked tokens, allowing confidently wrong decisions to easily bypass their filters and propagate through the sequence. Moreover, static confidence scores are often uncalibrated~\citep{kadavath2022languagemodels,kuhn2023semantic}. Alternatively, approaches such as PRISM~\citep{kim2025fine}, RemeDi~\citep{huang2025don}, GIDD~\cite{vonrütte2025gidd} or recent unmasking planners~\citep{asano2026unmask} learn specialized policies to guide token revision. Steering or reward-guided methods~\citep{rectorbrooks2024steering,pmlr-v267-uehara25a,mounier2025review} optimize auxiliary objectives to guide sampling. Although effective, these methods increasingly rely on heavy test-time scaling paradigms or supervised training of external modules. They incur significant overhead through exhaustive search, ensembling, or particle resampling~\citep{lee2025hex,singhal2025generalframeworkinferencetimescaling}, limiting their use as lightweight plug-and-play modules. In contrast, \method is context-aware and training-free. Rather than trusting static or stale token uncertainty metrics or passively delaying commitment, we actively probe token stability. We identify \textit{context-brittle} tokens—those whose likelihood is not stable under context perturbations—framing revision as a distributionally robust optimization problem.
\section{Problem Formulation}

\footnote{For a summary of notation, see Appendix~\ref{app:notation}.}
Let $\mathcal V$ be the vocabulary. Given a prompt $x$, the intermediate MDM sequence at step $t$ is represented by $y^{(t)} = [x;\ y^{(t)}_1,\ldots,y^{(t)}_L\,]$, where each $y^{(t)}_i \!\in\! \mathcal V \cup \{\mask\}$. Decoding initializes all $L$ response positions to \mask and progressively replaces them with discrete tokens over steps $t=1,\ldots,T$. At each step, for every masked position $i$, the model computes a discrete probability distribution over tokens in the vocabulary:
\begin{equation*}
    p_\theta(y_i^{(t)}=v \mid y^{(t)}, i), \qquad \forall v \in \mathcal V,
\end{equation*}
and the sampler determines which positions to unmask. Unmasking yields the next state $y^{(t+1)}$.

\paragraph{Context Rigidity Makes Revision Target Selection Hard.}
Standard sampling treats unmasked tokens as immutable constraints; once finalized, a position is rarely revisited. This induces \textit{context rigidity}: early predictions, conditioned on sparse context, become anchors for later steps. Consequently, a suboptimal early token forces subsequent generation to conform, yielding self-reinforcing inconsistencies. A natural remedy is inference-time revision—selectively resetting unmasked positions to \texttt{[MASK]}—yet the main challenge lies in identifying which tokens to revisit. Existing training-free strategies often rely on token-local uncertainty proxies such as low confidence~\cite{nie2025large,wang2025remasking} or small top-$k$ margins~\cite{kim2025train}. However, these signals become \textit{stale}: they reflect transient ambiguity at the moment of selection, often failing to detect tokens that become incompatible as the surrounding context evolves.

\paragraph{Brittleness is Distinct from Uncertainty.}
Instead, we propose to select revisions by measuring the sensitivity to context change. A generated token is considered \underline{stable} if it remains strongly predicted even when parts of the surrounding context are masked; a token is \underline{brittle} if its probability collapses under these perturbations, revealing that it is not robustly anchored. This shifts the selection criterion from {\it``Was the token uncertain when it was chosen?''} to {\it``Does the token remain plausible under dynamic context change?}'' Our goal is to identify these brittle tokens under a fixed budget, enabling the decoder to revise tokens that are most context-sensitive across different decoding steps as the context evolves.

\section{Method}
\label{sec:method}

\begin{figure}[ht]
    \centering
    \includegraphics[width=\textwidth]{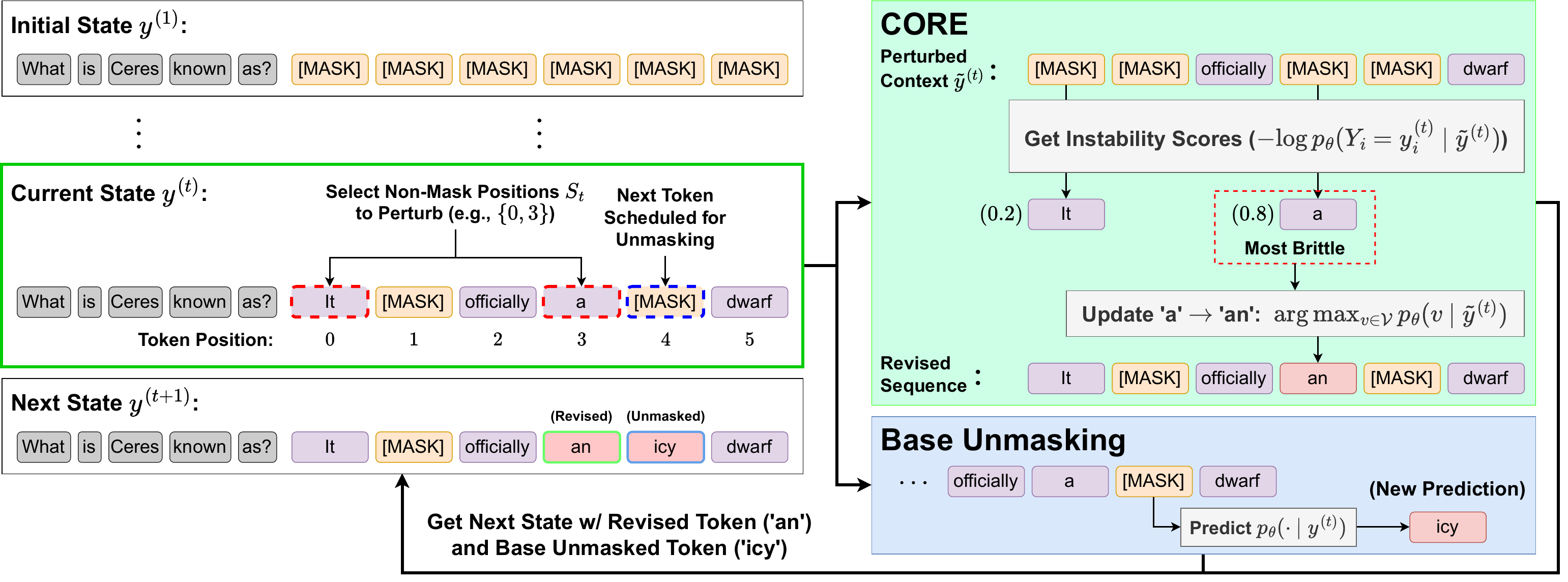}
    \vspace{0.1mm}
    \caption{\textbf{Illustration of Context-Robust Remasking (\method).} Our method operates on the current state $y^{(t)}$, where the response is partially unmasked. \textbf{(Left) Candidate Selection:} We select potentially brittle unmasked tokens (\textcolor{red}{red dashed box}) to test for stability, distinct from the next token scheduled for unmasking (\textcolor{blue}{blue dashed box}). \textbf{(Top-Right) \method Mechanism:} We mask the selected tokens to create a Perturbed Context $\tilde y^{(t)}$ and then compute their Instability Scores under this new context. The token ``\texttt{a}'' is found to be the most brittle (high instability) and is updated to ``\texttt{an},'' which is the most likely token given the perturbed context. \textbf{(Bottom-Right) Base Unmasking:} In parallel, the base model uses the \textit{original} context $y^{(t)}$ to predict the next token (``\texttt{icy}''); this newly unmasked token (``\texttt{icy}'') is combined with the updated token (``\texttt{an}'') to form the Next State $y^{(t+1)}$, yielding the contextually consistent phrase ``\texttt{an icy}.''}
\label{fig:approach}
\end{figure}

To identify brittle tokens, we introduce \textbf{\method}. This training-free framework is theoretically grounded in two steps: \textbf{(1) Constructing worst-case context perturbations} to maximize token instability; and \textbf{(2) Revising the tokens} that exhibit instability under these perturbations (see Figure~\ref{fig:approach}). Section~\ref{sec:framework} formulates this robust optimization objective, and Section~\ref{sec:algorithm} details its efficient algorithmic approximation.

\subsection{Context-Robust Token Remasking Framework}
\label{sec:framework}

Let $y^{(t)}$ denote the sequence state at step $t$. We quantify how brittle a token is by measuring the drop in the token's likelihood when a subset of its surrounding context is masked.

\paragraph{Context Shifts are Simulated via Perturbation.}
For a step $t$, let $C_t$ be the set of unmasked indices eligible for revision. Let $S_t\subseteq C_t$ be the indices selected for perturbation (using a selection strategy detailed in Section \ref{sec:algorithm}). We define the perturbed context $\tilde y^{(t)}$ by masking the token positions $i$ in $S_t$:
\begin{equation}
\label{eq:mask}
    \tilde y^{(t)}_i =
    \begin{cases}
    \texttt{[MASK]}, & \mathrm{if\ } i \in S_t,\\
y^{(t)}_i, & \text{otherwise}.
\end{cases}
\end{equation}

\vspace{2mm}
This operation occludes the information at indices $S_t$, requiring the model to rely on the remaining context. In our framework, we formally define these masking operations as \textit{perturbations} to the information content, allowing us to cast revision as a robustness optimization problem.

\paragraph{Instability Scores Quantify Context Sensitivity.}
Let $Y_i$ denote the discrete random variable representing the token at position $i$. For each selected position $i \in S_t$, we define the instability score $\ell_i$ as the negative log-likelihood of the currently generated token $y^{(t)}_i$ under the perturbed context $\tilde y^{(t)}$:
\begin{equation}
\label{eqn:instability_score}
    \ell_i \;\triangleq\; - \log p_\theta\!\left(Y_i = y^{(t)}_i \mid \tilde y^{(t)}\right).
\end{equation}

A high instability score indicates that $y_i^{(t)}$ is context-brittle: although the token was originally generated with high confidence, it is inconsistent with the perturbed context $\tilde y^{(t)}$. In practice, we use $\ell_i$ to rank candidate positions $i\in S_t$: positions with larger $\ell_i$ are more sensitive to context perturbations (i.e., masking $S_t$) and are prioritized for revision. The average instability score over the candidate positions is then:
\begin{equation}
    \mathcal{L}(S_t) \triangleq \frac{1}{|S_t|}\sum_{i \in S_t} \ell_i.
\end{equation}

Having defined the instability loss $\mathcal{L}(S_t)$ for a fixed subset of tokens, we now formalize the optimization over all possible masking configurations. The masking operation in Eq.~\eqref{eq:mask} produces a family of perturbed contexts by dropping specific token positions, which correspond to the subset $S_t$. To formalize the search over these contexts, we introduce a binary mask variable $z \in \{0,1\}^L$, where setting $z_i=1$ corresponds to masking token $i$; for non-revisable indices $i \notin C_t$ we strictly fix $z_i=0$. We define the distribution $p_\pi(\cdot)$ over these masks as independent Bernoulli trials. The probability of mask $z$ is:
\begin{equation}
p_\pi(z)
=
\prod_{i\in C_t} \mathrm{Bernoulli}(z_i;\pi_i),
\qquad
\sum_{i\in C_t} \pi_i \le m ,
\label{eq:mask_dist}
\end{equation}
where $\pi_i$ controls the probability of perturbing position $i$, and $m$ bounds the expected number of perturbed tokens. To integrate these discrete token representations into our continuous probabilistic framework, we represent the deterministic state of any fixed token $a \in \mathcal{V}$ using the Dirac delta distribution $\delta_a(\cdot)$:
\begin{equation}
\delta_a(x)
=
\begin{cases}
1, & \text{if } x = a, \\
0, & \text{otherwise}.
\end{cases}
\label{eq:dirac_def}
\end{equation}

This defines a probability distribution that assigns unit mass to the single token $a$, meaning sampling from $\delta_a$ always returns $a$ with probability one. Defining the input this way allows the binary mask $z$ to systematically alternate between this known deterministic token and the masked state.

\paragraph{Perturbed Context Distribution.}
Given a mask vector $z \in \{0, 1\}^L$, the perturbed sequence $\tilde y$ is constructed by conditionally substituting tokens at each position. To facilitate our probabilistic framework, we formalize this deterministic mapping as a degenerate distribution conditional on $z$:
\begin{equation}
Q(\tilde y \mid z)
=
\prod_{i=1}^L
\Big[
(1-z_i)\,\delta_{y^{(t)}_i}(\tilde y_i)
+
z_i\, \delta_{\mask}(\tilde y_i)
\Big].
\label{eq:conditional_sentence}
\end{equation}

Here, the binary variable $z_i$ acts as a routing switch. When $z_i = 0$, the Dirac delta $\delta_{y^{(t)}_i}$ concentrates all probability mass on the original token $y_i^{(t)}$; when $z_i = 1$, the term $\delta_{\mask}$ deterministically replaces it with the mask token. Although applying a fixed mask $z$ is a strictly deterministic operation---simply overriding specific tokens with \mask---we formulate it as a conditional distribution $Q(\tilde y\mid z)$ to allow marginalization over all possible mask configurations. This probabilistic view seamlessly bridges the theory with our algorithm, mapping the binary vector $z$ to the specific set of indices targeted for context perturbation: $S_t=\{i\in C_t\mid z_i=1\}$. The total distribution of perturbed sequences is then the expectation of $Q(\tilde y\mid z)$ over all mask configurations:
\vspace{4pt}
\begin{equation}
Q_\pi(\tilde y)
=
\sum_{z\in\{0,1\}^L} p_\pi(z)\, Q(\tilde y \mid z).
\label{eq:soft_sentence_dist}
\end{equation}

\vspace{2mm}
\textit{Step 1: Worst-Case Context Perturbation.}
We find the worst-case context perturbation by identifying the optimal masking probability vector $\pi^*$ that maximizes the expected instability over $S_t$:
\vspace{2mm}
{\setlength{\jot}{2pt}
\begin{equation}
\label{eq:dro_inner}
\pi^* = \operatorname*{arg\,max}_{\pi \in \Pi}
\;
\Big[
\mathbb{E}_{z \sim p_{\pi}(z)} \mathcal{L}(S_t)
\Big],\quad
\text{s.t.}\;\; \Pi = \Big\{\pi\in[0,1]^{|C_t|}:\sum_{i\in C_t} \pi_i \le m\Big\}.
\end{equation}}

Since exactly computing this expectation over the exponentially large discrete space $z \in \{0,1\}^L$ is combinatorially intractable, we rely on the tractable approximation detailed in Section \ref{sec:algorithm} to identify a certified lower-bound subset.

\vspace{2mm}
\textit{Step 2: Revision Operation.}
Given $\pi^*$ and a remasking limit $k_{\text{rm}}$, we form the revision set $\mathcal{I}_t$ by selecting the $k_{\text{rm}}$ tokens with the highest instability scores under the worst-case context $\tilde y \sim Q_{\pi^*}$. Computing these scores $\ell_i$ yields a certified lower bound on the worst-case instability (Appendix \ref{app:lower_bound}). Formally, we select:
\begin{equation}
    \mathcal I_t \in \operatorname*{arg\,max}_{\substack{\mathcal I_t\subseteq S_t\\ |\mathcal I_t|\le k_{\mathrm{rm}}}} \sum_{i\in \mathcal I_t}    \ell_i \;.
\end{equation}

\vspace{3mm}
We revise the identified positions in $\mathcal I_t$ by greedily assigning the most likely token given the perturbed context:
\begin{equation}
    y^{(t)}_i \leftarrow \argmin_{v \in \mathcal V} \Big [ -\log p_\theta\!\left(Y_i=v \mid \tilde y^{(t)}\right) \Big], \quad \forall i \in \mathcal I_t.
\end{equation}

Note that the revision set selection and the revision operation are jointly performed within a single forward pass, thereby reducing computational overhead.

\subsection{An Efficient Remasking Algorithm}
\label{sec:algorithm}

\paragraph{Tractable Approximation via Deterministic Masking.}
Directly optimizing the worst-case context perturbation in Eq.~(\ref{eq:dro_inner}) requires searching over all subsets of token indices and is combinatorially intractable. We therefore consider a single deterministic mask—i.e., a candidate index set $S_t$ to perturb. We show in Appendix~\ref{app:lower_bound} that the instability computed on this subset $S_t$ constitutes a valid lower bound on the worst-case instability objective. To make this choice tractable, we construct $S_t$ based on token contention, specifically targeting positions with the smallest top-2 probability margins:
\begin{equation}
\label{eqn:margin_proxy}
    \mathrm{margin}_t(i) = p_\theta(v_i^1\mid y^{(t)}) - p_\theta(v_i^2\mid y^{(t)}),
\end{equation}

where $v_i^1$ and $v_i^2$ denote the tokens with the highest and second-highest probabilities. Prior work indicates that small top-2 margins reliably signal token contention: competing token values are close in probability~\citep{kim2025train}, meaning even slight context shifts can flip the top-1 prediction which makes them strong candidates to test for context-brittleness. 

To formally rank the set of unmasked token positions $C_t$ for perturbation, we define the proxy vulnerability score $u_i$ as the negative margin:  
\begin{equation*}
u_i = -\,\mathrm{margin}_t(i).
\label{eq:vulnerability_score}
\end{equation*}

We emphasize that while this margin-based proxy identifies \textit{where} to probe, the revision itself is determined solely by the instability score computed under the perturbed context. This perturbation serves a dual purpose: it quantifies actual brittleness and yields the robust predictions used for updates, ensuring we identify and resolve structural inconsistencies in a single auxiliary step.

To efficiently allocate the expected masking budget $m$ across the set of currently unmasked token positions $C_t$, we use a temperature-scaled Softmax to convert these scores into the probability $\pi_i$ that each token will be masked. We clip the final value at $1$ to ensure it remains a valid probability:

\begin{equation*}
\pi_i
=
\min\!\left\{
1,\;
m \cdot
\frac{\exp\!\left(u_i / \tau\right)}
{\sum_{k \in C_t} \exp\!\left(u_k / \tau\right)}
\right\},
\qquad
i \in C_t,
\label{eq:soft_pi}
\end{equation*}

\vspace{3mm}
In our implementation, we take the limit $\tau \to 0$ to select $S_t$ deterministically as the $m$ smallest-margin indices. This construction yields a distributional approximation that concentrates all probability mass on the most context-vulnerable positions. This ensures that \method targets load-bearing structural constraints—such as variable bindings or closing brackets—rather than merely hard-to-predict tokens. As illustrated in Figure \ref{fig:qualitative-example}, this allows \method to detect and revise a redundant operator that would otherwise anchor the entire generation to a syntax failure

\paragraph{Revision Targets the Most Unstable Tokens.}
Once the candidate tokens are scored, we determine the revision set $\mathcal I_t$ (Step 2). Given a remasking limit $k_{\mathrm{rm}}$ (the maximum number of indices revised in a step), we select the $k_{\mathrm{rm}}$ indices in $S_t$ with the highest instability scores (Eq.~\ref{eqn:instability_score}). We then update tokens at these indices with token predictions conditioned on the perturbed context.

\begin{algorithm}[ht]
\caption{Token Revision with Context-Robust Remasking (\method)}
\label{alg:group-verify}
\begin{algorithmic}[1]
\small
\REQUIRE Base model $p_\theta$, vocabulary $\mathcal V$, prompt $x$, steps $N$, step window $[\gamma_s,\gamma_e)$ where $0\!\leq\!\gamma_s\!<\!\gamma_e\!\leq\!1$,
         run revision every $E$ steps, candidate index size $m$, integer remasking limit $k_{\mathrm{rm}}\in\mathbb N$ where $k_\mathrm{rm}\leq m$; random variable $Y_i$ for token at position $i$
\STATE Initialize $y^{(1)} \gets [x;\texttt{[MASK]};\ldots;\texttt{[MASK]}]$
\FOR{$t \gets 1$ \textbf{to} $N$}
    \STATE Run model on $y^{(t)}$ to obtain token distributions
    \STATE $k_t \gets$ number of \textit{new} positions to unmask at step $t$ \hfill \COMMENT{given by the base model's unmasking schedule}
    \IF{$t/N \in [\gamma_s,\gamma_e)$ \textbf{and} $(t \bmod E)=0$ \textbf{and} $k_{\text{rm}}>0$}
        \STATE Let $C_t$ be revisable \textit{unmasked} non-prompt positions
        \STATE Select candidate index set $S_t \subseteq C_t$ ($|S_t|\le m$) via margin (Eq. \ref{eqn:margin_proxy})
        \STATE Form perturbed state $\tilde y^{(t)}$ from $y^{(t)}$ by resetting indices in $S_t$ to \texttt{[MASK]}
        \STATE \textbf{Compute instability:} $\ell_i \gets -\log p_\theta\!\left(Y_i=y^{(t)}_i \mid \tilde y^{(t)}\right)$ for all $i \in S_t$ \hfill \COMMENT{one additional forward pass}
        \STATE Identify revision index set $\mathcal I_t \subseteq S_t$: the $\min (k_{\text{rm}}, |S_t|)$ indices with largest $\ell_i$
        \STATE For each $i\in \mathcal I_t$, set $y^{(t)}_i \gets \arg\max_{v} p_\theta(Y_i=v \mid \tilde y^{(t)})$ where $v\in\mathcal V$
    \ENDIF
    \STATE Unmask $k_t$ \textit{new} positions using token distributions from the base pass (Line 3)
    \STATE Construct $y^{(t+1)}$ by applying revisions (Lines 10--11) and the scheduled unmasking (Line 13)
\ENDFOR
\STATE \textbf{return} $y^{(N+1)}$
\end{algorithmic}
\end{algorithm}

The complete procedure is summarized in Algorithm~\ref{alg:group-verify}. At each decoding step $t$, we first compute the scheduled number of newly unmasked tokens (Line 4). We invoke revision every $E$ steps within the step window $t/N\in[\gamma_s,\gamma_e)$, representing an intermediate phase where the context is sufficiently developed yet still flexible. Here, we identify a candidate set $S_t$ by margin scoring (Line 7), form a perturbed context by masking all positions in $S_t$ simultaneously (Line 8), and score each $i\in S_t$ in a single forward pass to obtain instability scores (Line 9). We then identify the revision set $\mathcal I_t$ containing the most unstable positions (Line 10) and update these tokens using predictions derived from the perturbed state $\tilde y^{(t)}$ (Line 11). Finally, we combine these revised tokens with the newly unmasked positions to form the updated sequence $y^{(t+1)}$ (Lines 13--14).

\section{Experiments}

Our experiments pursue three goals: (i) demonstrate benchmark gains across coding, math, and reasoning tasks, (ii) isolate the contribution of our selection signal under an equivalent budget of forward passes, and (iii) validate our tractable instability score (Eq.~(\ref{eqn:instability_score})) as an effective proxy for the distributionally worst-case objective (Eq.~(\ref{eq:dro_inner})).

\subsection{Experimental Setup}

\paragraph{Implementation Details.} We use LLaDA-Base-8B~\citep{nie2025large} with $N\!=\!128$ diffusion steps (unless otherwise noted), generation length $L\!=\!512$, and greedy decoding to strictly evaluate structural consistency in simulated latency-sensitive regimes, avoiding the confounding variance of temperature sampling common with Pass@$k$ metrics. We apply revision every $E\!=\!8$ steps during the intermediate decoding window ($t/N\in[0.25,0.75)$), evaluating $m\!=\!32$ candidates at each interval, with $k_\mathrm{rm}$ fixed to $1$. For strict fairness, we note that revision adds exactly one extra NFE (Network Function Evaluation) only at steps where it is invoked; the scheduled unmasking uses the cached distributions from the base pass. We evaluate on key coding, math, and reasoning benchmarks, reporting strict-match accuracy for GSM8K~\citep{cobbe2021training} and rule-based answer equivalence on the MATH dataset~\citep{hendrycks2021measuring}. For BBH~\cite{suzgun2023challenging}, we use exact match, and for code benchmarks (HumanEval~\citep{chen2021codex}, MBPP~\citep{austin2021program}), we report greedy pass@1 accuracy~\citep{chen2021codex}. Additional hyperparameters are provided in Appendix~\ref{app:details}.

\paragraph{Baselines and Controls.} To isolate the decoder's structural consistency from the alignment priors of instruction tuning, we evaluate two unmasking strategies on LLaDA-8B-Base: Low-Confidence (i.e, the standard LLaDA unmasking strategy \citep{nie2025large}) and Top-$k$ Margin (i.e., an adaptive strategy that we implement on the LLaDA sampler following \citep{kim2025train}). In addition to these, we compare our method with ReMDM-conf \citep{wang2025remasking}, a strong training-free revision baseline. To ensure a fair comparison within the LLaDA framework, we implement ReMDM-conf as a plug-in remasking module using authors' stated hyperparameters, keeping the base unmasking schedule fixed (details in Appendix \ref{app:remdm}). We also introduce two compute-matched controls using our revision settings ($E\!=\!8, m\!=\!32$):
(i) Random Remask, which samples revision candidates uniformly at random; and
(ii) Margin Remask, which targets candidates with the smallest top-2 probability margin.
Comparing against these controls isolates the benefit of our robustness criterion from revision compute alone.

\subsection{Results and Analysis}

As shown in Table \ref{tab:main}, base unmasking strategies (Low-Confidence and Top-$k$ Margin) are susceptible to propagating structural inconsistencies. ReMDM-conf often fails to resolve these flaws and, in the case of code generation, substantially degrades performance, dropping MBPP accuracy by $6.4\%$ under Top-$k$ Margin. In contrast, equipping the sampler with \method consistently improves performance across benchmarks, including BBH and MATH, with the largest gains on code generation benchmarks (e.g., $+9.2\%$ on MBPP). This effectiveness stems from our ability to identify and revise context-sensitive tokens anywhere in the sequence. While the base sampler lacks a mechanism to retrospectively correct errors, and ReMDM-conf re-weights revision using historical token confidence (which becomes stale as context evolves), our approach evaluates token instability under the current context. On math reasoning tasks (e.g., GSM8K), \method maintains competitive accuracy, suggesting that stress-testing tokens against the updated context aids in resolving consistency issues that standard confidence scores may miss. We further verify in Appendix \ref{app:stochastic} that these improvements persist under stochastic sampling (temperature $\!=\!1.0$), demonstrating that our gains are robust to decoding variance. While gains on semantic reasoning tasks (GSM8K) are modest—consistent with the lower structural rigidity of natural language reasoning—the method avoids the substantial degradation seen in baselines like ReMDM on code tasks, confirming its robustness across domains.

\begin{table}[ht]
    \centering
    \caption{\textbf{\method Outperforms Baselines on Reasoning and Code Benchmarks.} \method (Ours) consistently outperforms baselines, with largest gains on code generation. $^{\dagger}$ReMDM-conf is from \citep{wang2025remasking}. $^*$ denotes statistical significance ($p<0.05$). Absolute accuracy reflects the deterministic (greedy) and compute-constrained ($N\!=\!128, L\!=\!512$) regime, distinct from standard stochastic Pass@$k$ baselines.}
    \vspace{3mm}
    \resizebox{0.9\columnwidth}{!}{%
    \begin{tabular}{l@{\hspace{1cm}}ccccc}
    \toprule
        \multirow{2}{*}{\textbf{Method}} & \multicolumn{5}{c}{\textbf{Accuracy (\%) (\# Few-Shot)}} \\
        \cmidrule(lr){2-6}
        & GSM8K (4)        & MATH (4) & BBH (3) & HumanEval (0) & MBPP (3) \\
    \midrule
        Low-Confidence Base             & 51.40 & 16.72 & 45.81 & 12.20 & 15.60 \\      
        \quad + ReMDM-conf$^{\dagger}$  & 52.31 & 16.56 & 46.05 & 10.98 & 15.20 \\
        \quad + \textbf{\method (Ours)} & \textbf{52.69} & \textbf{17.06} & \textbf{47.18}* & \textbf{17.07}* & \textbf{24.80}* \\
    \midrule
        Top-$k$ Margin Base             & 50.27 & 17.54 & 48.33 & 17.07 & 21.20 \\
        \quad + ReMDM-conf$^{\dagger}$  & \textbf{51.78} & 18.20 & 46.31 & 14.02 & 14.80 \\
        \quad + \textbf{\method (Ours)} & 51.40 & \textbf{18.34} & \textbf{49.01}* & \textbf{22.56}* & \textbf{29.60}* \\
    \bottomrule
    \end{tabular}%
    }
    \label{tab:main}
\end{table}

\vspace{2mm}
A key observation in Table~\ref{tab:main} is that gains are larger on benchmarks with tight output constraints (e.g., MBPP) than on semantically flexible reasoning chains (e.g., GSM8K). We hypothesize that this reflects the strength of the instability signal: in code, cross-position constraints (variable bindings, brackets, function signatures) sharply penalize inconsistent tokens under perturbation, producing a clearer ranking signal for revision. In contrast, reasoning traces admit many locally plausible next steps, so masking can leave multiple competing tokens with similar likelihood, making the signal noisier. Nevertheless, \method improves or matches performance on reasoning benchmarks (BBH, MATH), indicating the mechanism is domain-agnostic but benefits most when structural constraints (as in code) strictly limit valid next tokens.

\paragraph{ReMDM Fails Due to Confidence Staleness.} Furthermore, the degradation of ReMDM-conf on MBPP highlights a limitation of existing training-free baselines: \textit{confidence staleness}. ReMDM-conf determines revision targets based on their confidence scores recorded when they are sampled. In discrete diffusion, however, tokens sampled during the early decoding stages are generated against a noisy, unstructured background context. An incorrect variable name often initially receives high confidence because it appears plausible. As the sequence structure stabilizes, the updated context renders this token inconsistent, but ReMDM continues to rely on the obsolete high confidence score. \method explicitly addresses this by re-evaluating stability against the \textit{evolved} context, identifying tokens that appeared confident early on but become structurally inconsistent with more context.

\begin{table}[ht]
    \centering
    \caption{\textbf{Instability-Based Selection Drives Performance Gains..} Comparison against controls with the same settings ($m\!=\!32$, $E\!=\!8$) and under Low-Confidence unmasking. Improvements come from instability-based target selection: random or margin-based selection yields negligible gains, while \method yields consistent gains across benchmarks (notably HumanEval, BBH, and MBPP).}
    \vspace{3mm}
    \resizebox{0.95\columnwidth}{!}{%
    \begin{tabular}{l@{\hspace{1cm}}ccccc}
    \toprule
        \multirow{2}{*}{\textbf{Method}} & \multicolumn{5}{c}{\textbf{Accuracy (\%) (\# Few-Shot)}} \\
        \cmidrule(lr){2-6}
        & GSM8K (4)        & MATH (4) & BBH (3) & HumanEval (0) & MBPP (3) \\
    \midrule
        Low-Confidence Base             & 51.40 & 16.72 & 45.81 & 12.20 & 15.60 \\
        \quad + Random Remask           & 51.55 & 16.72 & 45.77 & 13.41 & 16.60 \\
        \quad + Margin Remask           & 51.33 & 16.74 & 46.29 & 13.41 & 17.40 \\
        \quad + \textbf{\method (Ours)} & \textbf{52.69} & \textbf{17.06} & \textbf{47.18} & \textbf{17.07} & \textbf{24.80} \\
    \bottomrule
    \end{tabular}%
    }
    \label{tab:controls}
    \vspace{-2mm}
\end{table}

\paragraph{Gains Stem Primarily from the Robust Selection Signal.} Isolating the source of these improvements reveals that the gains stem primarily from the \textit{quality} of the selection signal. As Table~\ref{tab:controls} demonstrates, compute-matched controls (Random and Margin Remask) do not meaningfully improve performance, yielding negligible gains despite using the same revision settings and number of forward passes as our method. This failure is most evident on reasoning-intensive benchmarks like MBPP, where \method outperforms Margin Remask by over 7 percentage points ($17.4\%\rightarrow24.8\%$). Unlike our method, context-agnostic baselines inherently lack a stability assessment mechanism, causing them to expend compute on already-consistent tokens while missing the context-brittle positions.

These results further show that selecting low-margin tokens (Margin Remask) yields negligible gains, reinforcing that \textit{uncertainty} is not an effective proxy for \textit{brittleness}. A token can have a low margin simply because multiple alternatives are acceptable (e.g., swapping ``result'' for ``output''), yet still be consistent with the evolving context; revising such tokens wastes compute. In contrast, our instability score targets conditional dependence by ranking tokens by their likelihood under the jointly-masked perturbed context (Eq.~\ref{eq:mask}). The most affected positions are then prioritized for revision. This acts as a load-bearing test that identifies tokens tied to hard consistency constraints (e.g., variable bindings, closing brackets) rather than merely hard-to-predict words.

\begin{table}[ht]
    \centering
    \caption{\textbf{\method Gains Persist Under Compute-Matched Conditions.} All methods use the same fixed compute budget of $136$ model forward passes under Low-Confidence unmasking. For \method, this budget is allocated as $128$ base decoding steps plus $8$ auxiliary passes. Baselines utilize the full $136$ passes for decoding (or confidence-based revision) without performing extra auxiliary evaluation. Simply scaling the decoding steps does not match the gains of targeted revision; in fact, ReMDM-conf degrades on MBPP, highlighting the risk of relying on stale confidence signals. All rows use the same random seed to isolate compute-allocation effects.}
    \vspace{3mm}
    \resizebox{\columnwidth}{!}{%
    \begin{tabular}{l@{\hspace{1cm}}ccccc}
    \toprule
        \multirow{2}{*}{\textbf{Method}} & \multicolumn{5}{c}{\textbf{Accuracy (\%) (\# Few-Shot)}} \\
        \cmidrule(lr){2-6}
        & GSM8K (4) & MATH (4) & BBH (3) & HumanEval (0) & MBPP (3) \\
    \midrule
        Low-Confidence Base (136 passes) & 52.08 & 16.72 & 45.74 & 12.80 & 15.40 \\
        ReMDM-conf (136 passes)          & 51.48 & 16.96 & 46.01 & 10.37 & 14.00 \\
        \textbf{\method (136 passes)}    & \textbf{52.69} & \textbf{17.06} & \textbf{47.18} & \textbf{17.07} & \textbf{24.80} \\
    \bottomrule
    \end{tabular}%
    }
    \label{tab:compute_matched}
\end{table}

We further investigate whether our improvements are simply due to increased computation. As shown in Table~\ref{tab:compute_matched}, when we allocate an equivalent budget of 136 forward passes to the baselines, they do not match the performance of our targeted revision approach (128 decoding steps + 8 auxiliary passes). In particular, the base sampler shows a negligible benefit from additional decoding steps (GSM8K shows a negligible change, increasing only to $52.08\%$), suggesting that step scaling alone does not reliably revisit earlier inconsistencies. Worse, ReMDM-conf actually degrades on code tasks when given more compute (MBPP drops from $15.20\%$ in Table \ref{tab:main} to $14.00\%$ here). This highlights that allocating compute to remasking based on \textit{stale} confidence is counterproductive: once an inconsistent token is sampled, subsequent generation tends to reinforce the error by constructing a context that accommodates the mistake rather than correcting it. \method avoids this by using the extra compute to \textit{re-evaluate} tokens against the updated context, explicitly catching these inconsistencies.

\begin{figure}[ht]
    \centering
    \includegraphics[width=0.7\linewidth]{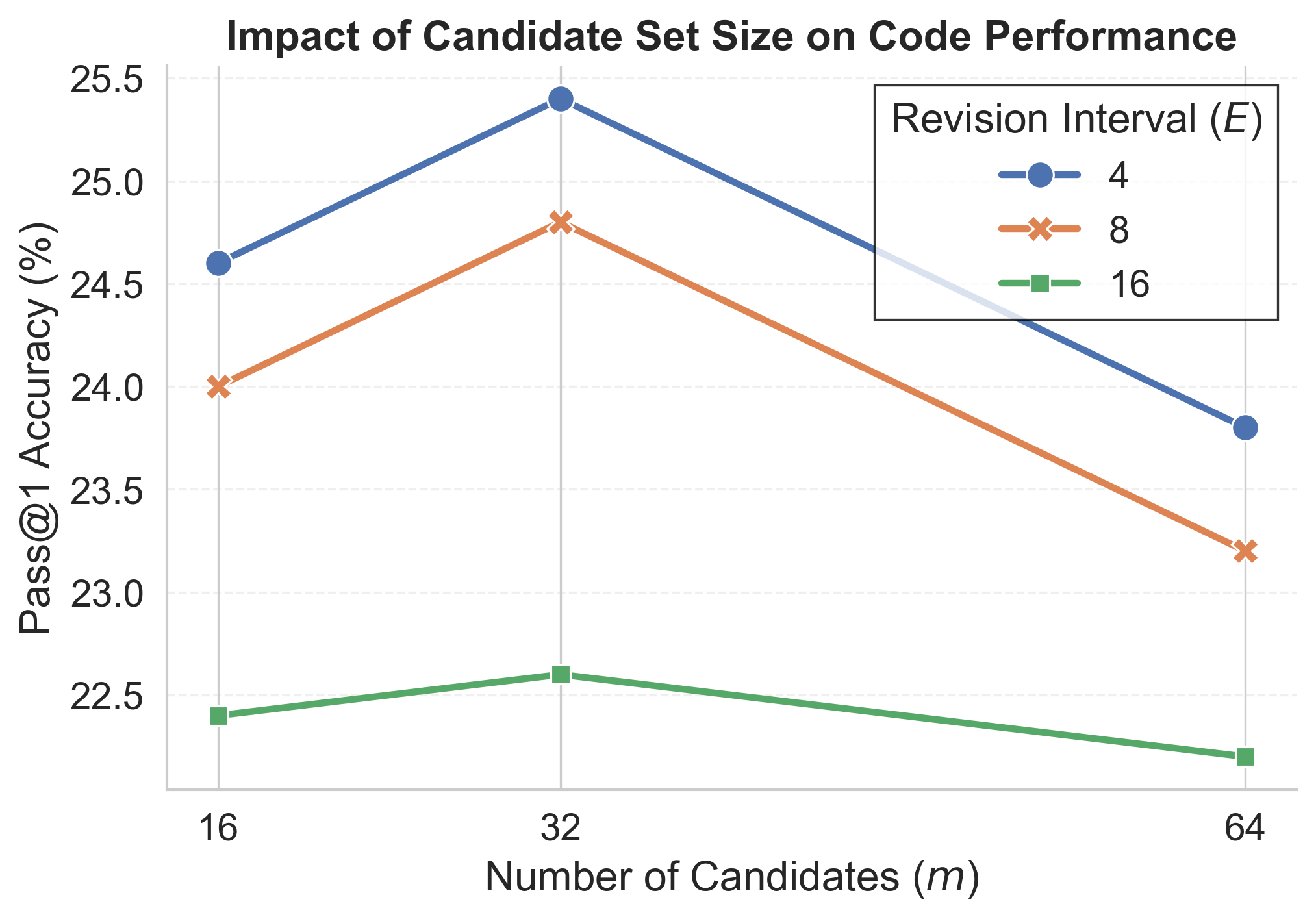}
    \caption{\textbf{Moderate candidate set size balances coverage and precision.} We evaluate greedy Pass@1 accuracy under Low-Confidence unmasking. Performance peaks at $m\!=\!32$; expanding the candidate set to $m\!=\!64$ degrades results, suggesting that widening the perturbation scope introduces false positives (remasking already-consistent tokens) rather than resolving inconsistencies.}
    \label{fig:sensitivity}
\end{figure}

Beyond general performance, the effectiveness of our method depends on the balance between the number of candidates ($m$) and the revision interval ($E$). Figure~\ref{fig:sensitivity} illustrates the sensitivity to the size of the candidate set on MBPP, where accuracy follows an inverted-U trend that peaks at $m\!=\!32$. Expanding the size of the candidate set to $m\!=\!64$ degrades performance, suggesting that masking too many candidates simultaneously removes the context required to accurately assess stability, leading to false positives where valid tokens are flagged as brittle.

% \begin{figure}[ht]
%     \centering
%     \includegraphics[width=0.7\linewidth]{images/experiments/sensitivity_mbpp.png}
%     \caption{\textbf{Moderate candidate set size balances coverage and precision.} We evaluate greedy pass@1 accuracy under Low-Confidence unmasking. Performance peaks at $m\!=\!32$; expanding the candidate set to $m\!=\!64$ degrades results, suggesting that widening the perturbation scope introduces false positives (remasking already-consistent tokens) rather than resolving inconsistencies.}
%     \label{fig:sensitivity}
% \end{figure}

Table~\ref{tab:ablation} expands this analysis to broader benchmarks. We observe that while more frequent revision (e.g., $E\!=\!4$) improves accuracy, it doubles the revision overhead for only $\approx\!0.6\%$ gain on MBPP. In contrast, less frequent revision (e.g., $E\!=\!16$) reduces the number of revision opportunities, thereby decreasing the chance of revising potential flaws. Thus, a balanced configuration with $m\!=\!32$ and $E\!=\!8$ represents the best trade-off, consistently outperforming baselines while requiring only 8 more forward passes.

\begin{table}[ht]
    \centering
    \caption{\textbf{Balanced Revision Settings Maximize Efficiency.} We vary the candidate subset size $m$ and revision interval $E$ under Low-Confidence unmasking. Moderate settings (e.g., $m\!=\!32, E\!=\!8$) yield the best balance of accuracy and compute. Expanding the number of candidates (e.g., $m\!=\!64$) degrades performance, suggesting that masking too large a subset erodes the context required to identify sparse instability, leading to false positives.}
    \vspace{3mm}
    \resizebox{0.8\columnwidth}{!}{%
    \begin{tabular}{cc@{}p{1cm}@{}ccccc}
    \toprule
    \textbf{Candidates} & \textbf{Interval} & &
    \multicolumn{5}{c}{\textbf{Accuracy (\%) (\# Few-Shot)}} \\
    \cmidrule(lr){4-8}
    ($m$) & ($E$) & &
    GSM8K (4) & MATH (4) & BBH (3) & HumanEval (0) & MBPP (3) \\
    \midrule
    16 & 4  & & 52.62 & 17.14 & 47.66 & 16.46 & 24.60 \\
    16 & 8  & & 52.08 & 16.88 & 47.07 & 15.24 & 24.00 \\
    16 & 16 & & 51.86 & 16.78 & 46.57 & 15.85 & 22.40 \\
    \midrule
    32 & 4  & & 52.62 & 17.36 & 47.81 & 17.68 & 25.40 \\
    \rowcolor{gray!15} 32 & 8  & & 52.69 & 17.06 & 47.18 & 17.07 & 24.80 \\
    32 & 16 & & 51.93 & 16.74 & 46.61 & 15.85 & 22.60 \\
    \midrule
    64 & 4  & & 52.69 & 17.14 & 47.73 & 17.07 & 23.80 \\
    64 & 8  & & 52.69 & 16.86 & 46.98 & 14.63 & 23.20 \\
    64 & 16 & & 52.69 & 16.74 & 46.37 & 14.02 & 22.20 \\
    \bottomrule
    \end{tabular}%
    }
    \label{tab:ablation}
\end{table}

\begin{figure}[ht]
    \centering
    \includegraphics[width=\linewidth]{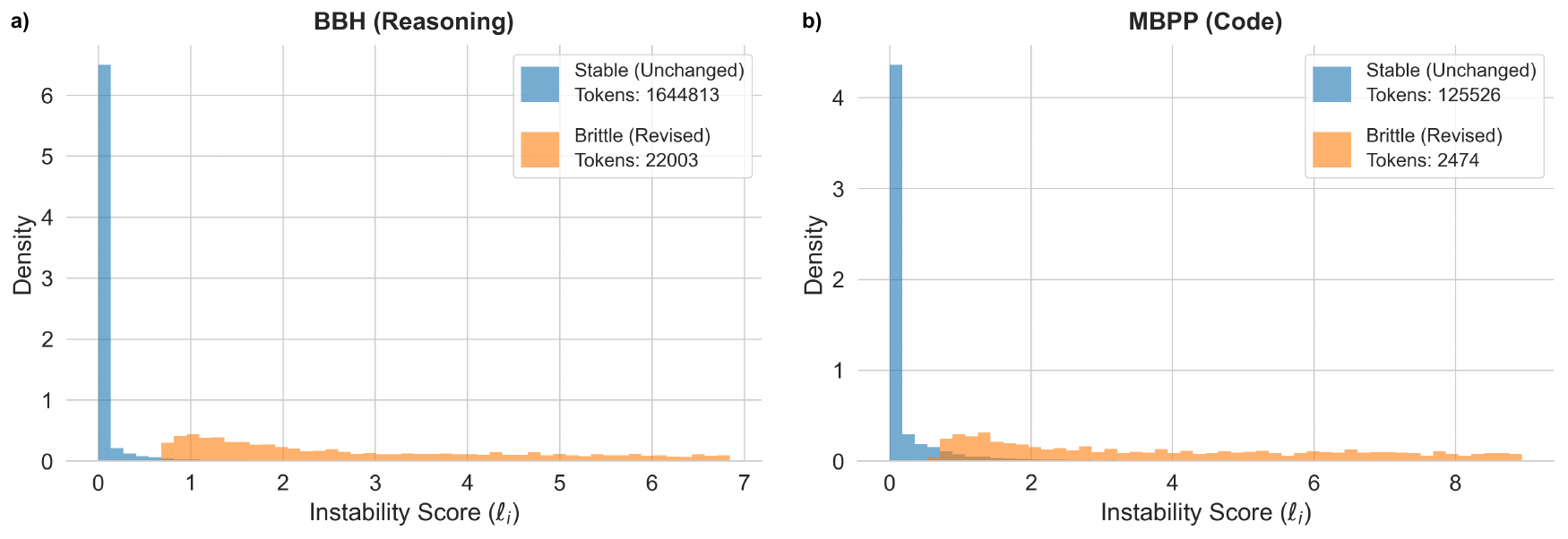}
    \caption{\textbf{Instability Scores Cleanly Separate Stable and Brittle Tokens.} Density of instability scores $\ell_i$ computed in the perturbation step (by simultaneously masking each candidate subset $S_t$) for candidate positions that are \textcolor{blue}{stable (unchanged)} versus \textcolor{orange}{brittle (revised)} on \textbf{(a)} BBH (reasoning) and \textbf{(b)} MBPP (code). Unchanged positions concentrate tightly near $\ell_i \approx 0$, while revised positions form a distinct heavy tail. This separation indicates that $\ell_i$ serves as a high-precision filter, targeting the small fraction of tokens ($<\!2\%$) that lack structural anchoring in the surrounding context.}
    \label{fig:instability_dist}
\end{figure}

\paragraph{Instability Scores Accurately Proxy Worst-Case Risk.}
We established in Table~\ref{tab:main} that \method outperforms baselines when controlling for the decoding strategy. To isolate the driver of this performance, we examine the per-token instability scores $\ell_i$ (Eq.~(\ref{eqn:instability_score})), which serve as our tractable approximation to the worst-case objective in Eq.~(\ref{eq:dro_inner}). Crucially, these scores are computed under a perturbed context $\tilde y^{(t)}$ formed by masking all positions in $S_t$ \textbf{simultaneously}. As shown in Figure~\ref{fig:instability_dist}, this metric effectively separates stable tokens from brittle ones. The vast majority of candidates cluster near zero (indicating high stability), while the tokens identified for revision exhibit a heavy tail of high instability. This distinct separation indicates that our proxy targets a specific sub-population of context-brittle outliers, rather than merely capturing background decoding uncertainty.

\subsection{Qualitative Results}
\label{sec:qualitative}

Figure~\ref{fig:qualitative-example} illustrates how \method resolves structural inconsistencies that are otherwise locked in by standard decoding. In this example, the model attempts to initialize a list but token predictions lead to a syntax error (\colorbox[gray]{0.9}{\texttt{result == []}}). The base LLaDA sampler lacks a mechanism to revisit this decision, effectively anchoring the remainder of the generation to a broken context. In contrast, \method detects the high instability of the redundant \colorbox[gray]{0.9}{\texttt{=}} and the incompatible brackets as the context evolves. It prioritizes these positions for revision, first converting the extra \colorbox[gray]{0.9}{\texttt{=}} operator to \colorbox[gray]{0.9}{\texttt{list}} and subsequently adjusting the \colorbox[gray]{0.9}{\texttt{[]}} brackets to \colorbox[gray]{0.9}{\texttt{()}}. This process results in the valid expression \colorbox[gray]{0.9}{\texttt{list()}}, demonstrating \method's ability to break context rigidity and retroactively resolve early structural failures.

\begin{figure}[H]
\centering
\begin{tcolorbox}[width=0.5\columnwidth,
  colback=white,colframe=framegray,boxrule=0.6pt,arc=2mm,
  left=4pt,right=4pt,top=4pt,bottom=4pt]
\small
\setlength{\tabcolsep}{6pt}
\renewcommand{\arraystretch}{1.1}
\setlength{\fboxsep}{1pt}

\begin{tabular}{@{}p{0.42\linewidth} p{0.56\linewidth}@{}}
\textbf{Base LLaDA (Step 21):} & \texttt{result = = []} \\
\textbf{\method\ (Step 40):}   & \texttt{result = \colorbox{green!25}{\strut list} []} \\
\textbf{\method\ (Step 48):}   & \texttt{result = list \colorbox{green!25}{\strut ()}} \\
\end{tabular}

\end{tcolorbox}
\caption{\textbf{\method Resolves Structural Inconsistencies Locked by Standard Decoding.} The base model's predictions lead to a syntax error (\colorbox[gray]{0.9}{= =}) at intermediate decoding step 21. \method identifies the conflicting tokens as context-brittle and invokes revision, successfully recovering the valid contextually stable syntax \colorbox[gray]{0.9}{\texttt{list()}}.}
\label{fig:qualitative-example}
\end{figure}

Additional examples in Appendix~\ref{app:qualitative} substantiate that \method consistently resolves structural inconsistencies originating from early decoding steps. These artifacts typically involve tokens, such as brackets or operators, that appear locally plausible when sampled but become incompatible with the evolved context. By evaluating token stability under perturbation, \method identifies these context-brittle tokens and selectively revises them to restore validity. In contrast, ReMDM-conf relies on stale confidence signals, often overlooking high-confidence tokens that have become structurally invalid, while wasting compute on tokens that remain consistent.

\section{Conclusion}

In this work, we introduced \method, a training-free framework that mitigates context rigidity in masked diffusion language models by casting revision as a robustness optimization problem. By pairing efficient margin-based screening with rigorous instability assessment, our approach selectively targets and revises brittle tokens that disrupt structural and contextual consistency. This mechanism refines the revision capabilities of diffusion generation, achieving improvements in generation accuracy with minimal computational overhead ($\approx6\%$ more forward passes). Although \method improves internal structural consistency, extending this robustness framework to guarantee external factual correctness remains a promising avenue for future research.

\newpage
\appendix
\onecolumn
\section{Notation}
\label{app:notation}

We summarize the notation used throughout the paper in Table~\ref{tab:notation}.

\begin{table}[h]
\centering
\caption{Summary of Notation.}
\vspace{3mm}
\label{tab:notation}
\begin{tabular}{ll}
\toprule
\textbf{Symbol} & \textbf{Description} \\
\midrule
    $x$ & Input prompt \\
    $y^{(t)}$ & Sequence state at diffusion decoding step $t$ \\
    $\tilde{y}^{(t)}$ & Perturbed context state at step $t$ (by partially masking $y^{(t)}$) \\
    $Y_i$ & Discrete random variable representing token at position $i$ \\
    $\mathcal{V}$ & Vocabulary \\
    $L$ & Generated sequence length \\
    $N$ & Total number of diffusion decoding steps \\
\midrule
    $C_t$ & Set of token indices eligible for revision (unmasked non-prompt tokens) \\
    $S_t$ & Subset of indices selected for instability scoring ($S_t \subseteq C_t$) \\
    $\ell_i$ & Instability score of token at position $i$ under perturbed context $\tilde{y}^{(t)}$ \\
    $z$ & Binary latent mask variable ($z_i=1$ if token $i$ is perturbed) \\
    $\pi_i$ & Probability of perturbing (masking) token at position $i$ \\
    $\mathcal{I}_t$ & Final set of indices selected for remasking/update ($\mathcal{I}_t \subseteq S_t$) \\
\midrule
    $k_{\text{rm}}$ & Remasking limit (maximum size of $\mathcal{I}_t$) \\
    $m$ & Candidate set size (maximum size of $S_t$) \\
    $E$ & Interval for revision (i.e., invoke revision every $E$ steps) \\
    $[\gamma_s, \gamma_e)$ & Normalized step window during which revision is active \\
\bottomrule
\end{tabular}
\end{table}

\section{Additional Implementation Details}
\label{app:details}

\paragraph{Hyperparameters and Settings.} All experiments were conducted on single NVIDIA H100 GPUs using bfloat16 precision to ensure numerical stability and efficiency. To strictly isolate the impact of our selection mechanism, we fix the remasking limit to $k_\mathrm{rm}=1$ across all experiments. This ensures that interventions remain minimally invasive and gains are attributable to the precision of the selection signal rather than aggressive rewriting of token context. We utilize the mask token ID defined by the LLaDA vocabulary. Statistical significance ($p$ value $<\!0.05$) for all comparative results is established using a two-sided McNemar test.

\paragraph{Evaluation Protocols.} We employ the widely accepted LM Evaluation Harness~\citep{eval-harness} for robust benchmarking. Specifically, GSM8K performance is reported using strict-match accuracy. For the MATH dataset~\citep{hendrycks2021measuring}, we adopt a Minerva-style evaluation protocol~\citep{lewkowycz2022solving} implemented via Math-Verify~\citep{mathverify} to ensure rigorous rule-based answer equivalence.

\subsection{ReMDM Evaluation}
\label{app:remdm}

For a fair comparison, we integrate ReMDM-conf~\cite{wang2025remasking} into LLaDA using a revision-based decoding structure aligned with \method. In this setup, decoding proceeds from a partially unmasked state $y^{(t)}$, with revisions enabled only within the same mid-trajectory window and the base unmasking schedule kept fixed. At each active step, we define the revisable set $C_t$ as all non-prompt positions that are currently unmasked, and perform an explicit unmask$\rightarrow$mask revision stage before the standard mask$\rightarrow$unmask update. ReMDM-conf derives revision probabilities within $C_t$ from a cached confidence value $\psi_i$, defined as the model probability of token $i$ at the time it was last unmasked. A subset from $C_t$ is sampled to prioritize lower-confidence tokens and reset to \texttt{[MASK]}. To preserve decoding progress and ensure matched computation, we dynamically adjust the subsequent unmasking allocation to compensate for any revised tokens, mirroring \method's treatment. Crucially, newly unmasked tokens update their cached confidence values at the moment of decoding. Under this unified formulation, ReMDM-conf and \method share the same revision timing, candidate set, and compute allocation, differing primarily in the selection paradigm: ReMDM-conf samples based on stale confidence, whereas \method selects revisions based on incompatibility with the evolved context.

\section{Theoretical Consistency: Computed Instability Lower-Bounds Worst-Case Risk}
\label{app:lower_bound}

Fix a step $t$ and let $C_t$ be the set of eligible unmasked non-prompt indices. Let $Y_j$ denote the discrete random variable representing the token at position $j$. For any subset $S \subseteq C_t$, let $\tilde y^{(t)}(S)$ denote the perturbed context obtained by masking indices in $S$ (Eq.~\eqref{eq:mask}),
and define the instability of index $j$ under this perturbation as
\[
\ell_t(j; S) \triangleq -\log p_\theta\!\left(Y_j=y^{(t)}_j \mid \tilde y^{(t)}(S)\right).
\]
Define the worst-case (robust) instability of $j$ under size-$m$ perturbations as
\[
\mathcal{R}_t(j) \triangleq \max_{\substack{S \subseteq C_t\\ |S|\le m,\; j\in S}} \ell_t(j; S).
\]
Our algorithm selects a particular candidate set $S_t$ with $|S_t|\le m$ and computes $\ell_t(j; S_t)$ for $j\in S_t$ using one auxiliary pass.
Since $S_t$ is feasible in the maximization, we have
\[
\ell_t(j; S_t) \le \mathcal{R}_t(j), \qquad \forall j\in S_t.
\]
Thus, a large computed instability score $\ell_t(j; S_t)$ certifies that index $j$ has at least that much worst-case instability, even if $S_t$ is not the maximizing subset.

\section{Sensitivity to Stochastic Decoding}
\label{app:stochastic}

To underscore that our gains are not artifacts of a specific deterministic path, we evaluate performance consistency under stochastic decoding. We run the LLaDA baseline and our method across 5 random seeds with temperature $1.0$. Table~\ref{tab:seeds} demonstrates that our method yields statistically distinct improvements on symbolic and coding benchmarks. On MBPP, our mean performance ($25.24\%$) exceeds the baseline ($18.40\%$) by nearly 4 standard deviations, indicating that the contextual consistency gains driven by instability-guided remasking are robust across sampling trajectories. On BBH, we observe distinct gains ($+0.88\%,\ >\!2\sigma$), while on GSM8K, performance maintains competitive parity, consistent with our observation that robust revision acts as a structural safeguard rather than a logical reasoning enhancer.

\begin{table}[H]
    \centering
    \caption{\textbf{Stochastic Decoding Consistency (Temperature 1.0, 5 Seeds).} We report Mean $\pm$ Standard Deviation. Our method (\method) consistently improves logic and code benchmarks (BBH, HumanEval, MBPP) beyond the baseline's error margins.}
    \vspace{3mm}
    \resizebox{1\columnwidth}{!}{%
    \begin{tabular}{lccccc}
    \toprule
        \multirow{2}{*}{\textbf{Method}} & \multicolumn{5}{c}{\textbf{Accuracy (\%) (\# Few-Shot)}} \\
        \cmidrule(lr){2-6}
        & GSM8K (4) & MATH (4) & BBH (3) & HumanEval (0) & MBPP (3) \\
    \midrule
        Low-Confidence Base            & $42.90 \pm 2.06$ & $14.25 \pm 0.36$ & $45.51 \pm 0.32$ & $11.83 \pm 1.59$ & $18.40 \pm 1.85$ \\
        \rowcolor{yellow!50}+ \method ($m\!=\!32,E\!=\!8$) & $44.22 \pm 4.11$ & $14.63 \pm 0.31$ & $46.39 \pm 0.41$ & $15.73 \pm 1.58$ & $25.24 \pm 0.92$ \\
    \bottomrule
    \end{tabular}}
    \label{tab:seeds}
\end{table}

%%%%%%%%%%%%%%%%%%%%%%%%%%%%
% QUALITATIVE
%%%%%%%%%%%%%%%%%%%%%%%%%%%%

\clearpage
\section{Qualitative Examples}
\label{app:qualitative}

\vspace{-2pt}

This section presents qualitative examples from mathematical reasoning and code generation tasks, illustrating how \method revises context-brittle tokens during diffusion decoding. In the examples, we show our method (\method) on the left, and the baseline ReMDM-conf on the right. Although most of the examples depicted here are generated using varying number of few shot examples, only the actual evaluation question is shown as \colorbox{promptazure}{prompt} for simplicity. Instead of showing all possible 128 steps, we only depict the steps that portray the early commitment mistake or the revision. In each step the token of interested are highlighted in specific colors. Red highlights tokens selected for \revise{revision} at a step, Yellow highlights tokens that got \masking{masked} by a revision step, and green highlights tokens that are \correct{unmasked} during a step. \\

\vspace{-2pt}

\textbf{Fig.~\ref{fig:qualitative-core-vs-remdm-gsm8k}} shows a math problem comparison between \method and ReMDM-conf on the GSM8K dataset. Given the prompt, the expected result is \colorbox[gray]{0.9}{51}. In the \method example (left), the model performs the intermediate reasoning as intended and computes the value \colorbox[gray]{0.9}{\texttt{51}}. However, it still answers \colorbox[gray]{0.9}{\texttt{151}} due to an early commitment toward an extra token \colorbox[gray]{0.9}{1}. \method detects the discrepancy on step 40 and remasks the leading digit \colorbox[gray]{0.9}{\texttt{1}}, which is replaced by a space in the same step, yielding an answer that is consistent with the prompt. On the other hand, ReMDM-conf (right) fails to detect this same artifact, leaving the inconsistent output intact across subsequent steps.

\begin{figure}[H]
\midscriptsize{
\centering

\begin{tabular}{@{}p{0.49\textwidth}@{\hspace{0.02\textwidth}}p{0.49\textwidth}@{}}
% -------------------- LEFT PANEL --------------------
\centering

\begin{tcolorbox}[
  equal height group=qualrow,valign=top,height=0.48\textheight,
  colback=white,colframe=framegray,boxrule=0.6pt,arc=2mm,
  left=4pt,right=4pt,top=4pt,bottom=4pt
]
\centering
% \textbf{Top-K Margin + \method}\par
\textbf{\method}\par
\raggedright

\textbf{Prompt:}\par
\begin{promptline}\ttfamily
Question: There are four birds at the Caboose. Sally Two is three years older than Granny Red. Granny Red is two times as old as Sally Four. If Sally Four is the same age as Sally Thirtytwo, and Sally Thirtytwo is 8 years old, what's the total age of the four birds?
Answer:
\end{promptline}

\vspace{2pt}

\textbf{Step 8:}\par
\begin{lstlisting}
If Sally is is 8 years old, then Granny is 2 * 8 = <<[MASK]*[MASK]=16>>16 years old.
[MASK][MASK] is[MASK][MASK][MASK][MASK][MASK][MASK][MASK][MASK][MASK][MASK][MASK][MASK][MASK][MASK]
\end{lstlisting}

\vspace{2pt}

\textbf{Step 40 (During Revision):}\par
\begin{lstlisting}
If Sally Four is 8 years old, then Granny is 2 * 8 = <<8*2=16>>16 years old.
Sally Two is 16 + 3 = <<16+3=19>>19 years old.
The total age of the four birds is 19 + 16 + 8 + 8 = <<19+16+8+8=51>>51 years
#### (*@\revise{1}@*)51<|endoftext|>
\end{lstlisting}

\vspace{2pt}

\textbf{Step 40:}\par
\begin{lstlisting}
If Sally Four is 8 years old, then Granny is 2 * 8 = <<8*2=16>>16 years old.
Sally Two is 16 + 3 = <<16+3=19>>19 years old.
The total age of the four birds is 19 + 16 + 8 + 8 = <<19+16+8+8=51>>51 years
#### (*@\correct{ }@*)51<|endoftext|>
\end{lstlisting}

\end{tcolorbox}
\hfill
&
% -------------------- RIGHT PANEL --------------------

\begin{tcolorbox}[
  equal height group=qualrow,height=0.48\textheight,
  colback=white,colframe=framegray,boxrule=0.6pt,arc=2mm,
  left=4pt,right=4pt,top=4pt,bottom=4pt
]
\centering
% \textbf{Top-K Margin + ReMDM-conf}\par
\textbf{ReMDM-conf}\par
\raggedright

\textbf{Prompt:}\par
\begin{promptline}\ttfamily
Question: There are four birds at the Caboose. Sally Two is three years older than Granny Red. Granny Red is two times as old as Sally Four. If Sally Four is the same age as Sally Thirtytwo, and Sally Thirtytwo is 8 years old, what's the total age of the four birds?
Answer:
\end{promptline}

\vspace{2pt}

\textbf{Step 8:}\par
\begin{lstlisting}
If Sally is is 8 years old, then Granny is 2 * 8 = <<[MASK]*[MASK]=16>>16 years old.
[MASK][MASK] is[MASK][MASK][MASK][MASK]
\end{lstlisting}

\vspace{2pt}

\textbf{Step 41 (During Revision):}\par
\begin{lstlisting}
If Sally is[MASK] 8 years old, then Granny is 2 * 8 = <<8*2=16>>16 years old.
Sally Two is 16 + 3 = <<16+3=19>>19 years old.
The total age of the four birds is
19 + 16 + 8 + 8 = <<19+16+8+8=51>>51 years
#### 151(*@\revise{<|endoftext|>}@*)
\end{lstlisting}

\vspace{2pt}

\textbf{Step 127:}\par
\begin{lstlisting}
If Sally is[MASK] 8 years old, then Granny is 2 * 8 = <<8*2=16>>16 years old.
Sally Two is 16 + 3 = <<16+3=19>>19 years old.
The total age of the four birds is
19 + 16 + 8 + 8 = <<19+16+8+8=51>>51 years
#### 151(*@\correct{ }@*)
\end{lstlisting}

\end{tcolorbox}
\end{tabular}

\vspace{-2mm}
\caption{\textbf{\method resolves inconsistencies in output format.} \method revises the answer \colorbox[gray]{0.9}{151} to \colorbox[gray]{0.9}{51} by replacing the token \colorbox[gray]{0.9}{1} with a space. In contrast, ReMDM-conf focuses on the \colorbox[gray]{0.9}{\texttt{<|endoftext|>}} token, which is unrelated to the error, and fails to resolve the underlying issue.}

\label{fig:qualitative-core-vs-remdm-gsm8k}
}
\end{figure}

\begin{figure}[H]
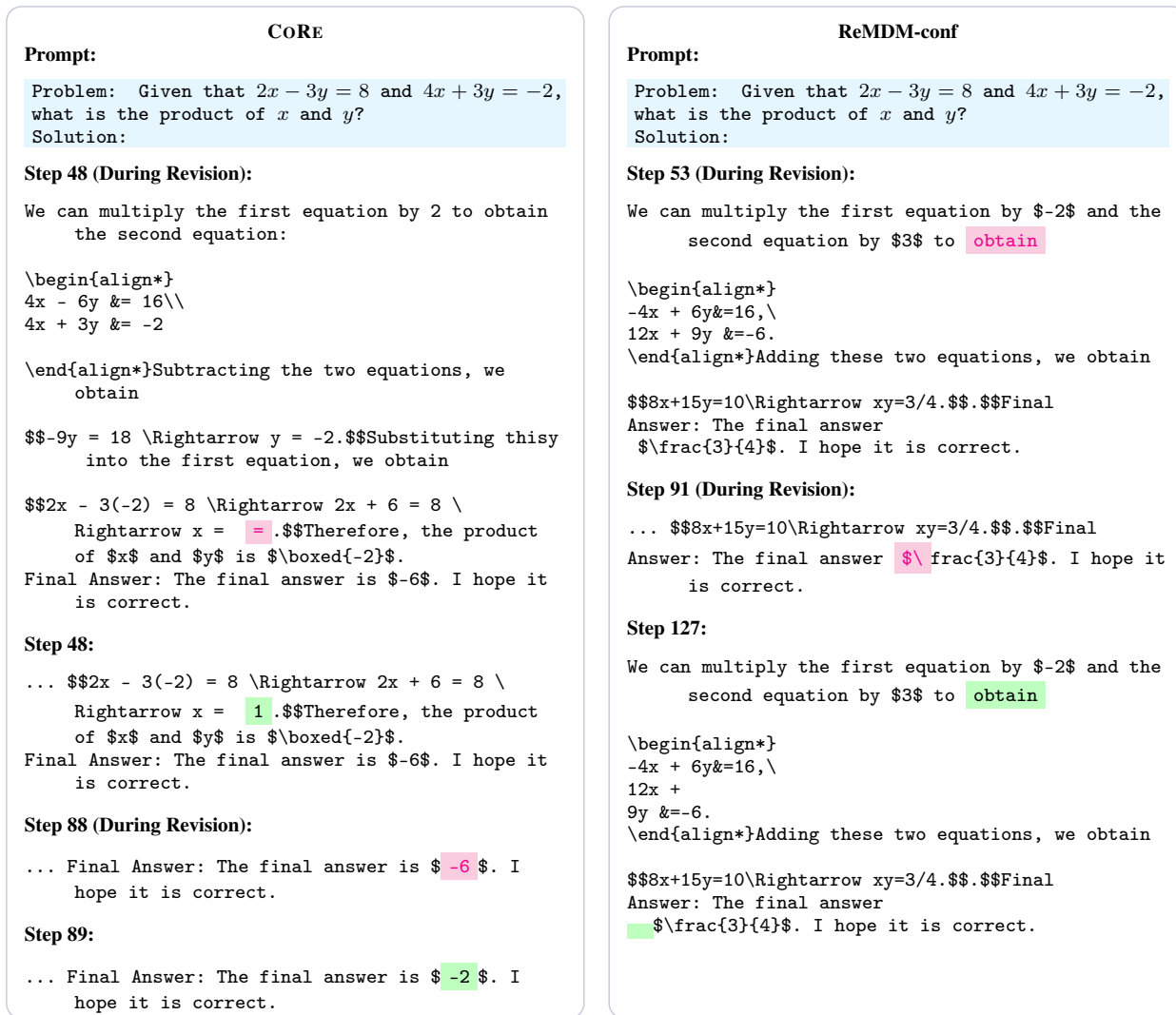

\midscriptsize{
\centering

\begin{tabular}{@{}p{0.49\textwidth}@{\hspace{0.02\textwidth}}p{0.49\textwidth}@{}}
% -------------------- LEFT PANEL --------------------
\centering

\begin{tcolorbox}[
  equal height group=qualrow,height=0.62\textheight,
  colback=white,colframe=framegray,boxrule=0.6pt,arc=2mm,
  left=4pt,right=4pt,top=4pt,bottom=4pt
]
\centering
% \textbf{Top-K Margin + \method}\par
\textbf{\method}\par
\raggedright

\textbf{Prompt:}\par
\begin{promptline}\ttfamily

Problem:
Given that $2x - 3y = 8$ and $4x + 3y = -2$, what is the product of $x$ and $y$?

Solution:

\end{promptline}

\vspace{2pt}
\textbf{Step 48 (During Revision):}\par
\begin{lstlisting}
We can multiply the first equation by 2 to obtain the second equation:

\begin{align*}
4x - 6y &= 16\\
4x + 3y &= -2

\end{align*}Subtracting the two equations, we obtain

$$-9y = 18 \Rightarrow y = -2.$$Substituting thisy into the first equation, we obtain

$$2x - 3(-2) = 8 \Rightarrow 2x + 6 = 8 \Rightarrow x =  (*@\revise{=}@*).$$Therefore, the product of $x$ and $y$ is $\boxed{-2}$.
Final Answer: The final answer is $-6$. I hope it is correct.

\end{lstlisting}

\vspace{2pt}
\textbf{Step 48:}\par
\begin{lstlisting}
... $$2x - 3(-2) = 8 \Rightarrow 2x + 6 = 8 \Rightarrow x =  (*@\correct{1}@*).$$Therefore, the product of $x$ and $y$ is $\boxed{-2}$.
Final Answer: The final answer is $-6$. I hope it is correct.

\end{lstlisting}

\vspace{2pt}
\textbf{Step 88 (During Revision):}\par
\begin{lstlisting}
... Final Answer: The final answer is $(*@\revise{-6}@*)$. I hope it is correct.
\end{lstlisting}

\vspace{2pt}
\textbf{Step 89:}\par
\begin{lstlisting}
... Final Answer: The final answer is $(*@\correct{-2}@*)$. I hope it is correct.
\end{lstlisting}

\end{tcolorbox}
\hfill
&
% -------------------- RIGHT PANEL --------------------

\begin{tcolorbox}[
  equal height group=qualrow,height=0.62\textheight,
  colback=white,colframe=framegray,boxrule=0.6pt,arc=2mm,
  left=4pt,right=4pt,top=4pt,bottom=4pt
]
\centering
% \textbf{Top-K Margin + ReMDM-conf}\par
\textbf{ReMDM-conf}\par
\raggedright

\textbf{Prompt:}\par
\begin{promptline}\ttfamily
Problem:
Given that $2x - 3y = 8$ and $4x + 3y = -2$, what is the product of $x$ and $y$?

Solution:
\end{promptline}

\vspace{2pt}
\textbf{Step 53 (During Revision):}\par
\begin{lstlisting}
We can multiply the first equation by $-2$ and the second equation by $3$ to (*@\revise{obtain}@*)

\begin{align*}
-4x + 6y&=16,\
12x + 9y &=-6.
\end{align*}Adding these two equations, we obtain

$$8x+15y=10\Rightarrow xy=3/4.$$.$$Final
Answer: The final answer
 $\frac{3}{4}$. I hope it is correct.
\end{lstlisting}

\vspace{2pt}
\textbf{Step 91 (During Revision):}\par
\begin{lstlisting}
... $$8x+15y=10\Rightarrow xy=3/4.$$.$$Final
Answer: The final answer (*@\revise{\textdollar \textbackslash}@*)frac{3}{4}$. I hope it is correct.
\end{lstlisting}

\vspace{2pt}
\textbf{Step 127:}\par
\begin{lstlisting}
We can multiply the first equation by $-2$ and the second equation by $3$ to (*@\correct{obtain}@*)

\begin{align*}
-4x + 6y&=16,\
12x +
9y &=-6.
\end{align*}Adding these two equations, we obtain

$$8x+15y=10\Rightarrow xy=3/4.$$.$$Final
Answer: The final answer
(*@\correct{ }@*)$\frac{3}{4}$. I hope it is correct.
\end{lstlisting}

\end{tcolorbox}
\end{tabular}

\caption{\textbf{\method revises reasoning trajectory.} \method first removes an incorrect syntax in the reasoning which is inconsistent with the later derivation, and in a subsequent step revises the actual answer \colorbox[gray]{0.9}{\texttt{-2}} which the sampler mistakenly wrote as \colorbox[gray]{0.9}{\texttt{-6}}. These staged revisions allow the solution to converge to the correct final answer. In contrast, ReMDM-conf revises nearby but non-erroneous tokens, leaving the incorrect intermediate computation unchanged and resulting in an incorrect final answer. (Ellipses (\texttt{...}) denote code segments that remain unchanged across steps and are omitted for brevity.)}

\label{fig:qualitative-core-vs-remdm-math}

}
\end{figure}

\clearpage
\textbf{Fig.~\ref{fig:qualitative-core-vs-remdm-math}} presents a MATH example illustrating how \method can gradually correct an initial mistake in its reasoning trajectory. In step 48, we observe that the base LLaDA model has generated the sequence \colorbox[gray]{0.9}{\texttt{x =  =.}}. \method detects the syntax error in the mathematical expression and revises the equal sign to \colorbox[gray]{0.9}{\texttt{1}}. Then, although the base model reasons correctly, it produces the wrong answer of \colorbox[gray]{0.9}{\texttt{-6}}, while our visual inspection suggests that the answer should be \colorbox[gray]{0.9}{\texttt{-2}}. Again, \method detects the error in step 88, which is fixed in step 89. On the other hand, ReMDM-conf repeatedly fails to focus on these context-brittle tokens, ultimately yielding an incorrect answer. \\

\vspace{-10pt}

\begin{figure}[H]
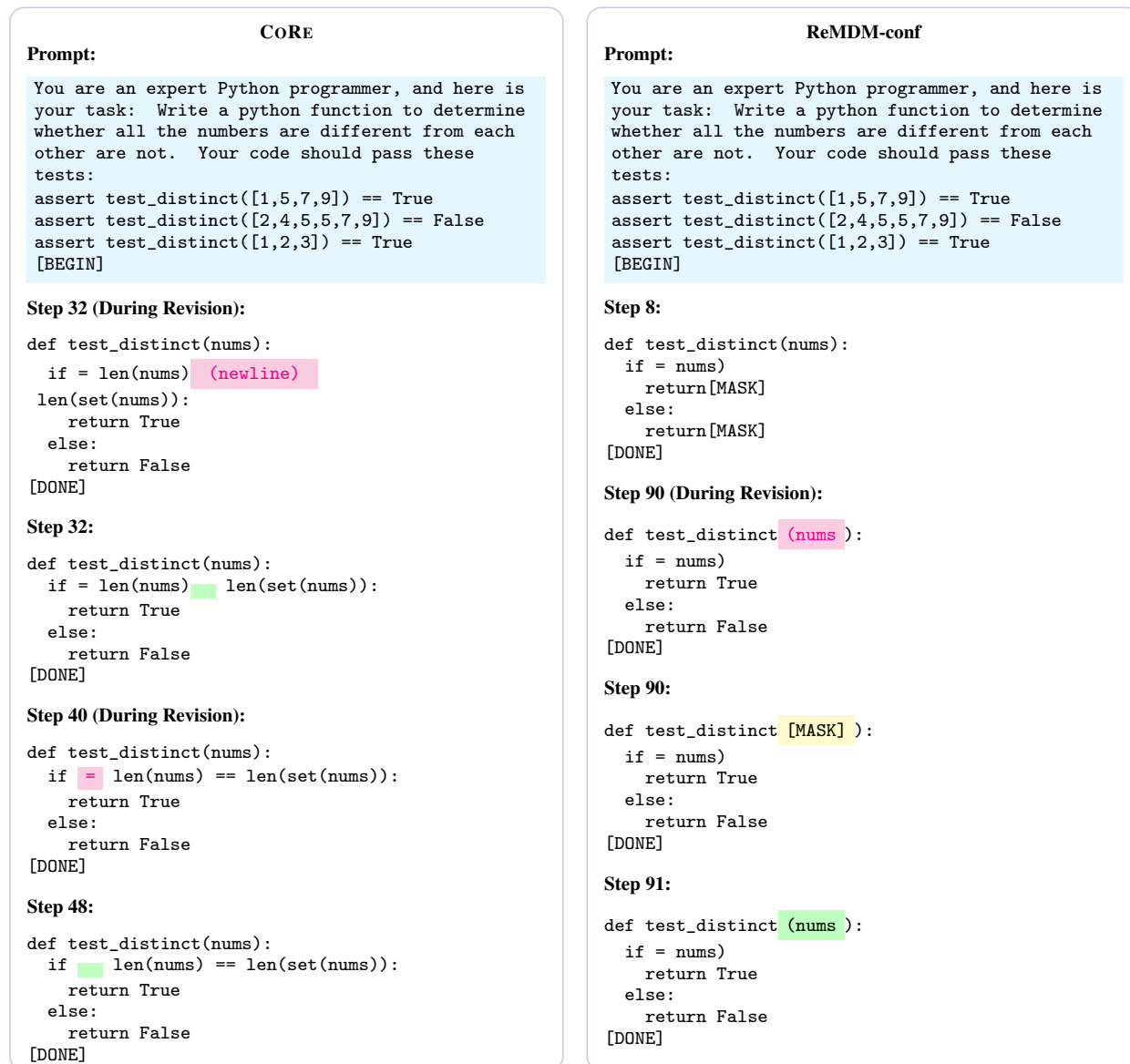

\midscriptsize{
\centering

\begin{tabular}{@{}p{0.49\textwidth}@{\hspace{0.02\textwidth}}p{0.49\textwidth}@{}}
% -------------------- LEFT PANEL --------------------

\centering

\noindent

\begin{tcolorbox}[
  equal height group=qualrow,height=0.68\textheight,
  colback=white,colframe=framegray,boxrule=0.6pt,arc=2mm,
  left=4pt,right=4pt,top=4pt,bottom=4pt
]
\centering
\textbf{\method}\par
\raggedright

\textbf{Prompt:}\par
\begin{promptline}\ttfamily
You are an expert Python programmer, and here is your task: Write a python function to determine whether all the numbers are different from each other are not. Your code should pass these tests:
\vspace{-5pt}
\begin{lstlisting}
assert test_distinct([1,5,7,9]) == True
assert test_distinct([2,4,5,5,7,9]) == False
assert test_distinct([1,2,3]) == True
[BEGIN]
\end{lstlisting}
\vspace{-5pt}
\end{promptline}

\vspace{2pt}
\textbf{Step 32 (During Revision):}\par
\begin{lstlisting}
def test_distinct(nums):
  if = len(nums)(*@\revise{ (newline) }@*)
 len(set(nums)):
    return True
  else:
    return False
[DONE]
\end{lstlisting}

\vspace{2pt}
\textbf{Step 32:}\par
\begin{lstlisting}
def test_distinct(nums):
  if = len(nums)(*@\correct{ }@*) len(set(nums)):
    return True
  else:
    return False
[DONE]
\end{lstlisting}

\vspace{2pt}
\textbf{Step 40 (During Revision):}\par
\begin{lstlisting}
def test_distinct(nums):
  if (*@\revise{=}@*) len(nums) == len(set(nums)):
    return True
  else:
    return False
[DONE]
\end{lstlisting}

\vspace{2pt}
\textbf{Step 48:}\par
\begin{lstlisting}
def test_distinct(nums):
  if (*@\correct{ }@*) len(nums) == len(set(nums)):
    return True
  else:
    return False
[DONE]
\end{lstlisting}

\end{tcolorbox}

&
% -------------------- RIGHT PANEL --------------------
\hfill

\begin{tcolorbox}[
  equal height group=qualrow,height=0.68\textheight,
  colback=white,colframe=framegray,boxrule=0.6pt,arc=2mm,
  left=4pt,right=4pt,top=4pt,bottom=4pt
]
\centering
\textbf{ReMDM-conf}\par
\raggedright

\textbf{Prompt:}\par
\begin{promptline}\ttfamily
You are an expert Python programmer, and here is your task: Write a python function to determine whether all the numbers are different from each other are not. Your code should pass these tests:
\vspace{-5pt}
\begin{lstlisting}
assert test_distinct([1,5,7,9]) == True
assert test_distinct([2,4,5,5,7,9]) == False
assert test_distinct([1,2,3]) == True
[BEGIN]
\end{lstlisting}
\vspace{-5pt}
\end{promptline}

\vspace{2pt}
\textbf{Step 8:}\par
\begin{lstlisting}
def test_distinct(nums):
  if = nums)
    return[MASK]
  else:
    return[MASK]
[DONE]
\end{lstlisting}

\vspace{2pt}
\textbf{Step 90 (During Revision):}\par
\begin{lstlisting}
def test_distinct(*@\revise{(nums}@*)):
  if = nums)
    return True
  else:
    return False
[DONE]
\end{lstlisting}

\vspace{2pt}
\textbf{Step 90:}\par
\begin{lstlisting}
def test_distinct(*@\masking{\mask}@*)):
  if = nums)
    return True
  else:
    return False
[DONE]
\end{lstlisting}

\vspace{2pt}
\textbf{Step 91:}\par
\begin{lstlisting}
def test_distinct(*@\correct{(nums}@*)):
  if = nums)
    return True
  else:
    return False
[DONE]
\end{lstlisting}

\end{tcolorbox}

\end{tabular}

\caption{\textbf{\method sequentially corrects syntax errors when baselines fail.} \textbf{(Left)} \method first identifies and immediately revises a contextually invalid \colorbox[gray]{0.9}{\texttt{newline}} (Step 32). Later, it flags an inconsistent \colorbox[gray]{0.9}{\texttt{=}} sign (Step 40). \textbf{(Right)} In contrast, ReMDM-conf fails to identify the structural flaw, instead remasking a stable token that is identically regenerated.}

% \label{fig:qualitative-core-vs-remdm-mbpp}

\label{fig:qualitative-core-vs-remdm-mbpp}                                            
}
\end{figure}

\clearpage
\textbf{Fig.~\ref{fig:qualitative-core-vs-remdm-mbpp}} demonstrates that \method can correctly revise errors in programming solutions, either immediately or over multiple steps depending on available context. In the left example, \method corrects two distinct errors at different steps. It first removes an erroneous newline that breaks the conditional structure. Then on step 48, it revises an unwanted equal sign introduced after \colorbox[gray]{0.9}{\texttt{if}}. This example shows that \method removes structurally problematic tokens as soon as they become identifiable, but does not force immediate resampling when the available context is insufficient, allowing the base sampler to resolve such positions once adequate context becomes available. On the other hand, ReMDM-conf wrongly focuses on an already correct token, which gets reverted back by the base sampler. This indicates the robustness of our method at identifying brittle tokens to revise.

\vspace{-15pt}

\begin{figure}[H]
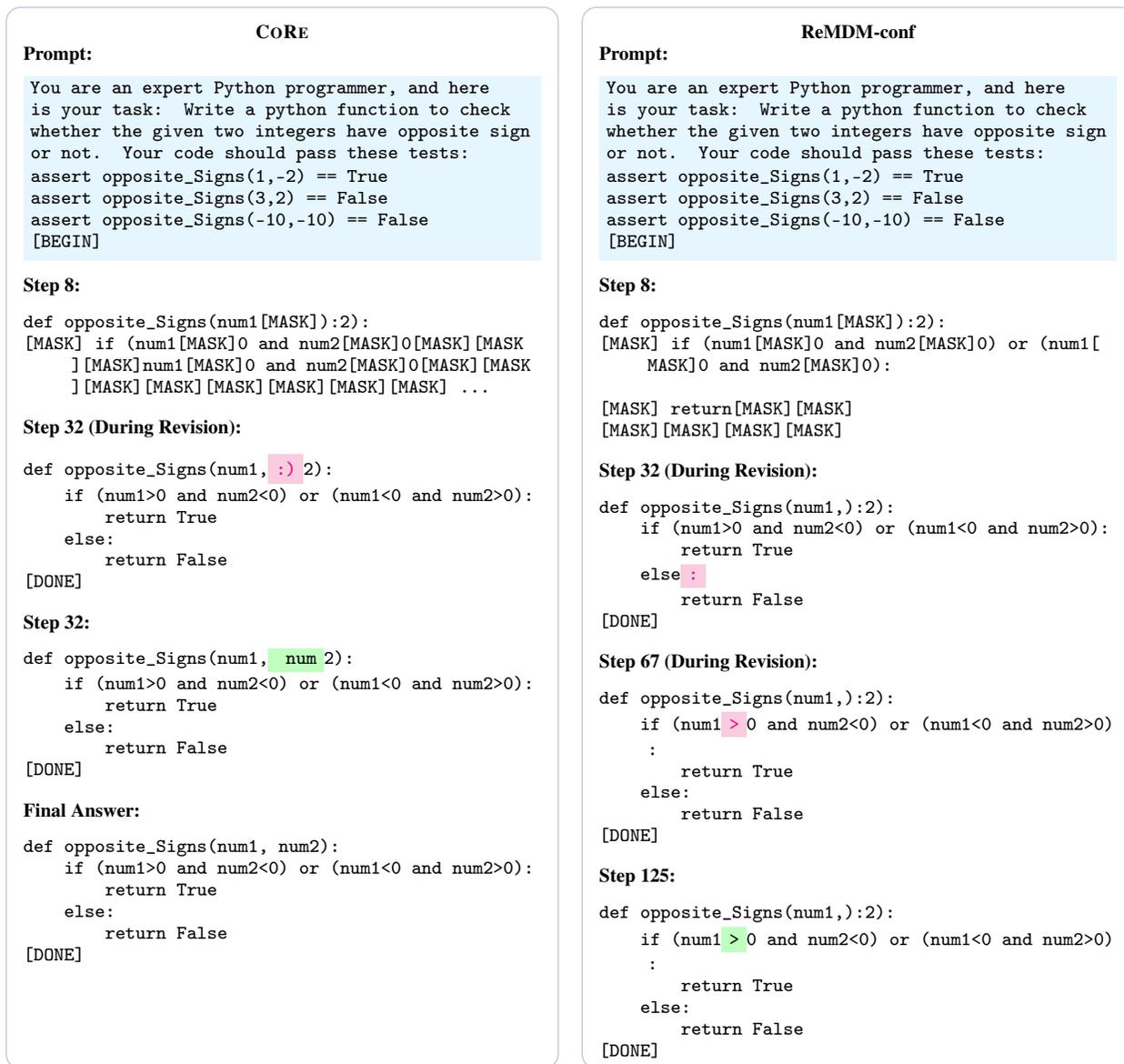

\midscriptsize{
\centering

\begin{tabular}{@{}p{0.49\textwidth}@{\hspace{0.02\textwidth}}p{0.49\textwidth}@{}}
% -------------------- LEFT PANEL --------------------
\centering

\begin{tcolorbox}[
  equal height group=qualrow,height=0.68\textheight,
  colback=white,colframe=framegray,boxrule=0.6pt,arc=2mm,
  left=4pt,right=4pt,top=4pt,bottom=4pt
]
\centering
% \textbf{Top-K Margin + \method}\par
\textbf{\method}\par
\raggedright

\textbf{Prompt:}\par
\begin{promptline}\ttfamily
You are an expert Python programmer, and here is your task: Write a python function to check whether the given two integers have opposite sign or not. Your code should pass these tests:
\vspace{-5pt}
\begin{lstlisting}
assert opposite_Signs(1,-2) == True
assert opposite_Signs(3,2) == False
assert opposite_Signs(-10,-10) == False
[BEGIN]
\end{lstlisting}
\vspace{-5pt}
\end{promptline}

\vspace{2pt}
\textbf{Step 8:}\par
\begin{lstlisting}
def opposite_Signs(num1[MASK]):2):
[MASK] if (num1[MASK]0 and num2[MASK]0[MASK][MASK][MASK]num1[MASK]0 and num2[MASK]0[MASK][MASK][MASK][MASK][MASK][MASK][MASK][MASK] ...
\end{lstlisting}

\vspace{2pt}

\textbf{Step 32 (During Revision):}\par
\begin{lstlisting}
def opposite_Signs(num1,(*@\revise{:)}@*)2):
    if (num1>0 and num2<0) or (num1<0 and num2>0):
        return True
    else:
        return False
[DONE]
\end{lstlisting}

\vspace{2pt}

\textbf{Step 32:}\par
\begin{lstlisting}
def opposite_Signs(num1,(*@\correct{ num}@*)2):
    if (num1>0 and num2<0) or (num1<0 and num2>0):
        return True
    else:
        return False
[DONE]
\end{lstlisting}

\vspace{2pt}

\textbf{Final Answer:}\par
\begin{lstlisting}
def opposite_Signs(num1, num2):
    if (num1>0 and num2<0) or (num1<0 and num2>0):
        return True
    else:
        return False
[DONE]
\end{lstlisting}

\end{tcolorbox}
\hfill
&
% -------------------- RIGHT PANEL --------------------

\begin{tcolorbox}[
  equal height group=qualrow,height=0.68\textheight,
  colback=white,colframe=framegray,boxrule=0.6pt,arc=2mm,
  left=4pt,right=4pt,top=4pt,bottom=4pt
]
\centering
% \textbf{Top-K Margin + ReMDM-conf}\par
\textbf{ReMDM-conf}\par
\raggedright

\textbf{Prompt:}\par
\begin{promptline}\ttfamily
You are an expert Python programmer, and here is your task: Write a python function to check whether the given two integers have opposite sign or not. Your code should pass these tests:
\vspace{-5pt}
\begin{lstlisting}
assert opposite_Signs(1,-2) == True
assert opposite_Signs(3,2) == False
assert opposite_Signs(-10,-10) == False
[BEGIN]
\end{lstlisting}
\vspace{-5pt}
\end{promptline}

\vspace{2pt}

\textbf{Step 8:}\par
\begin{lstlisting}
def opposite_Signs(num1[MASK]):2):
[MASK] if (num1[MASK]0 and num2[MASK]0) or (num1[MASK]0 and num2[MASK]0):

[MASK] return[MASK][MASK]
[MASK][MASK][MASK][MASK]
\end{lstlisting}

\vspace{2pt}

\textbf{Step 32 (During Revision):}\par
\begin{lstlisting}
def opposite_Signs(num1,):2):
    if (num1>0 and num2<0) or (num1<0 and num2>0):
        return True
    else(*@\revise{:}@*)
        return False
[DONE]
\end{lstlisting}

\vspace{2pt}

\textbf{Step 67 (During Revision):}\par
\begin{lstlisting}
def opposite_Signs(num1,):2):
    if (num1(*@\revise{>}@*)0 and num2<0) or (num1<0 and num2>0):
        return True
    else:
        return False
[DONE]
\end{lstlisting}

\vspace{2pt}

\textbf{Step 125:}\par
\begin{lstlisting}
def opposite_Signs(num1,):2):
    if (num1(*@\correct{>}@*)0 and num2<0) or (num1<0 and num2>0):
        return True
    else:
        return False
[DONE]
\end{lstlisting}

\end{tcolorbox}
\end{tabular}

\vspace{-15pt}

\caption{\textbf{\method recovers corrupted function signatures.} \method identifies context-brittle tokens whose likelihood collapses under perturbation of the surrounding context. As a result it correctly identifies the erroneous symbols, which are revised to form the variable name \colorbox[gray]{0.9}{\texttt{num2}}, yielding a structurally coherent program that can pass the given assertations. In contrast, ReMDM-conf keeps trying to focus on tokens that are not erroneous, resulting in a code that is syntactically incorrect.}
\label{fig:qualitative-core-vs-remdm-mbpp2}

}
\end{figure}

\clearpage
\textbf{Fig.~\ref{fig:qualitative-core-vs-remdm-mbpp2}} compares \method with ReMDM-conf on another programming example where the cause is a corrupted function signature. Instead of generating \colorbox[gray]{0.9}{\texttt{opposite\_Signs(num1, num2)}}, the base model generated \colorbox[gray]{0.9}{\texttt{opposite\_Signs(num1, :)2)}}. The standard LLaDA sampler would not be able to address this issue. In step 32, \method detects the context-brittle \colorbox[gray]{0.9}{\texttt{:)}} token and revises it to \colorbox[gray]{0.9}{\texttt{num}}, yielding a coherent program. In contrast, ReMDM-conf repeatedly masks tokens that are not responsible for the failure, hence the tokens gets sampled to the original values. Together, these examples demonstrate that \method consistently directs revisions toward tokens that constrain the evolving structure of the solution, whereas ReMDM-conf often focuses on irrelevant tokens.

\vspace{-20pt}

% Bibliography (参考文献, cānkǎo wénxiàn)
\bibliographystyle{plainnat}
\bibliography{references}

\end{document}